\documentclass[letterpaper]{article}

\usepackage{aaai2026}

\usepackage{times}
\usepackage{helvet}
\usepackage{courier}
\usepackage[hyphens]{url}
\usepackage{graphicx}
\usepackage{amsmath}
\usepackage{amssymb}
\usepackage{booktabs}

\usepackage{natbib}
\usepackage{algorithm}
\usepackage{algpseudocode}

\usepackage{verbatim}
\usepackage{microtype}

\usepackage{xcolor}

\title{Semantic Event Graphs for Long-Form Video Question Answering}

\author{
Aradhya Dixit\textsuperscript{\rm 1},
Tianxi Liang\textsuperscript{\rm 2}
}

\affiliations{
\textsuperscript{\rm 1}Wake Tech Community College\\
\textsuperscript{\rm 2}Cornell University\\
}

\begin{document}
\maketitle

\begin{abstract}
Long-form video question answering remains challenging for modern vision--language models, which struggle to reason over hour-scale footage without exceeding practical token and compute budgets. Existing systems typically downsample frames or feed dense visual embeddings to large-context language models, trading off temporal coverage against cost. We propose \textbf{Semantic Event Graphs} (SEG), a lightweight symbolic interface between video and language that replaces raw frames with compact temporal interaction logs. Our pipeline detects and tracks objects with YOLOv11, converts proximity patterns into \textit{START/END} human--object events, and organizes them into a \textbf{Temporal Scene Graph} (TSG). At inference time, a \textbf{query-aware pruning} module identifies anchor entities and lexically relevant events, returning only a small subgraph which is verbalized and passed to Gemini~2.5 Flash for answer generation. On five YouTube videos (300--500 interactions each) and 120 automatically generated long-horizon questions, SEG achieves \textbf{65.0\%} accuracy using only \textbf{3.47k} tokens per query, closely matching a full-log baseline (\textbf{62.5\%} at \textbf{40.39k} tokens) while reducing token usage by \textbf{91.4\%}. A short-context baseline restricted to the last 30 seconds collapses to \textbf{2.5\%} accuracy, underscoring the need for explicit temporal memory. These results show that symbolic temporal graphs can serve as an effective, plug-and-play memory layer for off-the-shelf vision--language models, preserving long-range reasoning ability while making long-form video question answering substantially more token- and cost-efficient. Code, logs, and event-extraction tools will be released for reproducibility.
\end{abstract}

\section{Introduction}

Understanding long-horizon videos is one of the most challenging open problems in
vision-language reasoning. A seemingly simple everyday recording---such as a
person preparing food or tidying a room---may span tens of minutes, contain
thousands of frames, and include a complex sequence of object interactions that
are semantically relevant only in sparse moments. A 10-minute video contains over
18{,}000 frames at 30\,FPS, yet most question-answering tasks reference only a
small subset of events. Directly feeding the entire stream into Vision-Language
Models (VLMs) results in excessive token usage and expensive inference. Even
state-of-the-art models such as \textit{Gemini 2.5 Flash} \cite{reid2024gemini},
\textit{GPT-4o} \cite{openai2024gpt4o}, and emergent video-language architectures
(e.g., Video-LLaMA \cite{li2023videollama}, LLaMA-VID \cite{ma2024llamavid},
OmniVL/OmniVLM \cite{ma2024omni}) excel at short-form understanding but degrade
sharply when extended beyond 30--300 seconds due to context window limits and
attention cost scaling \cite{kaplan2020scaling,openai2024o1}.

A fundamental observation motivates this work: \textbf{the world is not dense,
but event-driven}. Humans do not recall every frame when asked
``When did the person pick up the cup?'' but instead retrieve a structured mental
timeline of actions. Most reasoning requires ordering, duration, and causal
sequences, not raw pixels. Existing approaches to long-video QA attempt frame
sampling, hierarchical compression, or spatiotemporal embedding aggregation
\cite{bai2022egovlp,zhang2023vid2seq,kim2023videox}. However, these either lose
important interactions or scale poorly. What is missing is a representation that
\textbf{compresses videos semantically rather than visually}.

We address this gap by introducing \textbf{Semantic Event Graphs (SEG)}, a
lightweight intermediate structure that converts raw video frames into
human-object interaction events. Our system performs YOLOv11 detection and
persistent tracking \cite{jocher2023ultralytics}, identifies interactions via
centroid-based spatial proximity \cite{baradel2018object_interaction,qi2023egoschema},
and logs each interaction as a \texttt{START/END} event with timestamps. These
events are organized into a \textbf{Temporal Scene Graph (TSG)}, where nodes
represent entities and edges encode temporal relations, building on ideas from
temporal scene graphs and spatio-temporal relational models \cite{yuan2019tsg,tang2020sttran,ahuja2021noiselabels}.
When a natural language query is issued, a \textit{query-aware pruning} stage
selects only the events and entity subgraphs relevant to the question. The pruned
timeline is then transformed into a compact textual narrative passed to
\textbf{Gemini 2.5 Flash} for reasoning.

This pipeline enables reasoning over videos that are otherwise far too large to
process directly. Instead of thousands of frames, the model reasons over tens of
events---while still maintaining ordering, duration, and causal traceability.
Our evaluation across five Creative Commons YouTube videos shows that SEG reduces input tokens by \textbf{91.4\% on
average}, yet achieves \textbf{65.0\% QA accuracy}, matching full-context
LLM access while being \textit{12$\times$ cheaper} in token cost. Short-context
baselines collapse to \textbf{2.5\%}, demonstrating that long-term event memory
is essential for task success.

\textbf{Key idea:}
\vspace{-2mm}
\[
\begin{aligned}
\text{Raw Video} \rightarrow \text{Objects} \rightarrow \text{Events} \\
\rightarrow \text{Graph} \rightarrow \text{Pruned Timeline} \\
\rightarrow \text{LLM Answer}
\end{aligned}
\]
\vspace{-2mm}

\noindent Figure~\ref{fig:pipeline} depicts the workflow.

\subsection*{Contributions}

We provide:

1) \textbf{Semantic Event Graphs}, a symbolic compression framework for long-form
video reasoning, inspired by temporal scene graph and interaction-based activity
recognition \cite{yuan2019tsg,tang2020sttran,ahuja2021noiselabels}.\\
2) A \textbf{Temporal Scene Graph (TSG)} with query-seeded pruning for relevant
event retrieval, connected to retrieval-augmented and memory-enhanced LLMs
\cite{lewis2020rag,borgeaud2022retro,wu2024memorybank,fan2024longmem}.\\
3) A long-form Video-QA system using \textbf{Gemini 2.5 Flash} achieving
\textbf{91.4\% token reduction with maintained accuracy}, providing a token-efficient
alternative to dense video prompting \cite{deng2024video_token,li2023tokenmerging}.\\
4) A new dataset of five long-form YouTube videos with auto-generated
QA annotations, logs, and evaluation benchmarks, publicly released.\\

Together, these results indicate that structured symbolic representations can
serve as a scalable interface between continuous video streams and large-context
language models, opening the path towards long-horizon embodied reasoning without
GPU-heavy visual inference.

\section{Methods}

\begin{figure*}[t]
    \centering
    \includegraphics[width=\textwidth]{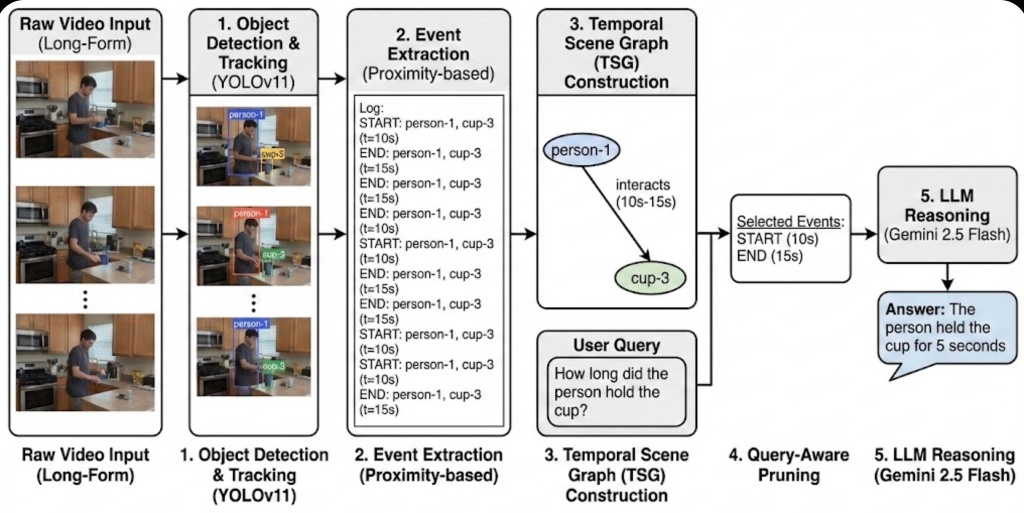}
    \caption{Method pipeline overview: raw video is processed by YOLOv11 detection and tracking, converted into START/END interaction events, assembled into a Temporal Scene Graph (TSG), pruned according to the query, and finally passed as a compact narrative to Gemini 2.5 Flash for question answering.}
    \label{fig:pipeline}
\end{figure*}

Our system is designed to transform raw long-form video into a compressed,
query-dependent event narrative suitable for efficient reasoning by large
language models. Figure~\ref{fig:pipeline} illustrates the full pipeline. The
method consists of four sequential components: (1) object detection and tracking,
(2) proximity-based event extraction, (3) Temporal Scene Graph construction, and
(4) query-aware pruning for token-efficient retrieval.

\vspace{2mm}
\subsection{Object Detection and Tracking}

We employ YOLOv11 (Ultralytics v8.3.235) for object detection with identity
persistence enabled via internal Re-ID tracking \cite{jocher2023ultralytics}.
For every frame $F_t$, YOLO produces a set of detections:
\[
det = \{id,\ label,\ bbox(x_1,y_1,x_2,y_2)\}
\]
where $id$ is the persistent track identity, $label$ represents the object class,
and $(x_1,y_1,x_2,y_2)$ denotes bounding box coordinates. We compute the centroid
of each detection:
\[
c_i = \Big( \frac{x_{1,i}+x_{2,i}}{2}, \frac{y_{1,i}+y_{2,i}}{2} \Big)
\]
Identity propagation ensures stable entity references across frames, which is
critical for interaction reasoning and aligns with prior work on object-centric
video modeling \cite{baradel2018object_interaction,qi2023egoschema}. In practice,
we prioritize GPU (CUDA) when available, fallback to MPS for Apple silicon, and
gracefully degrade to CPU when necessary. Tracking results are stored in
\texttt{video\_objects.json} and visualized in \texttt{tracked\_output.mp4}.

\vspace{2mm}
\subsection{Event Extraction via Proximity}

Once objects are tracked, we infer interactions based on spatial proximity. Two
detections $i$ and $j$ are considered interacting when:
\[
d_{ij} = \|c_i - c_j\|_2 \le \delta
\]
where $\delta$ is a configurable pixel threshold (default: 100px; 200px for
wide-angle scenes). We only record interactions involving humans through a
\texttt{focus\_class = "person"} filter, significantly reducing irrelevant noise,
similar to human-centric egocentric models \cite{bai2022egovlp,qi2023egoschema}.

If an inactive pair becomes close enough, a \texttt{START} event is triggered. If
either object disappears or moves apart for more than $\beta$ consecutive frames
(default: $\beta=5$), an \texttt{END} event is emitted. Each event is stored as a
structured dictionary:
\begin{verbatim}
{ "timestamp":t, "frame":f,
  "type":"START|END",
  "subject":"person-1", "object":"cup-3" }
\end{verbatim}
This conversion yields a compressed temporal log representing narrative structure
rather than raw pixels, reducing a 30-minute video from tens of thousands of
frames to typically 200--500 events.

\vspace{2mm}
\subsection{Temporal Scene Graph (TSG)}
\label{sec:tsg}

Within the SEG pipeline, we represent interactions as a directed multi-graph:
\[
\begin{aligned}
G &= (V, E), \\
V &= \{\text{unique subject/object IDs}\}, \quad 
E = \{\text{event edges}\}
\end{aligned}
\]
implemented using NetworkX \texttt{MultiDiGraph}. Each edge stores metadata:
\begin{itemize}
    \item event type (START/END)
    \item timestamp and frame index
    \item subject \& object identifiers
    \item raw JSON record for text conversion
\end{itemize}
This structure makes temporal reasoning explicit: nodes denote entities (e.g.,
\texttt{person-1}, \texttt{cup-3}), while edges represent temporal interactions.

Unlike visual embeddings, this graph is interpretable, symbolic, and memory-light,
and relates to prior temporal scene graph approaches \cite{yuan2019tsg,tang2020sttran,ahuja2021noiselabels}.

\vspace{2mm}
\begin{figure}[t]
  \centering
  \includegraphics[width=\linewidth]{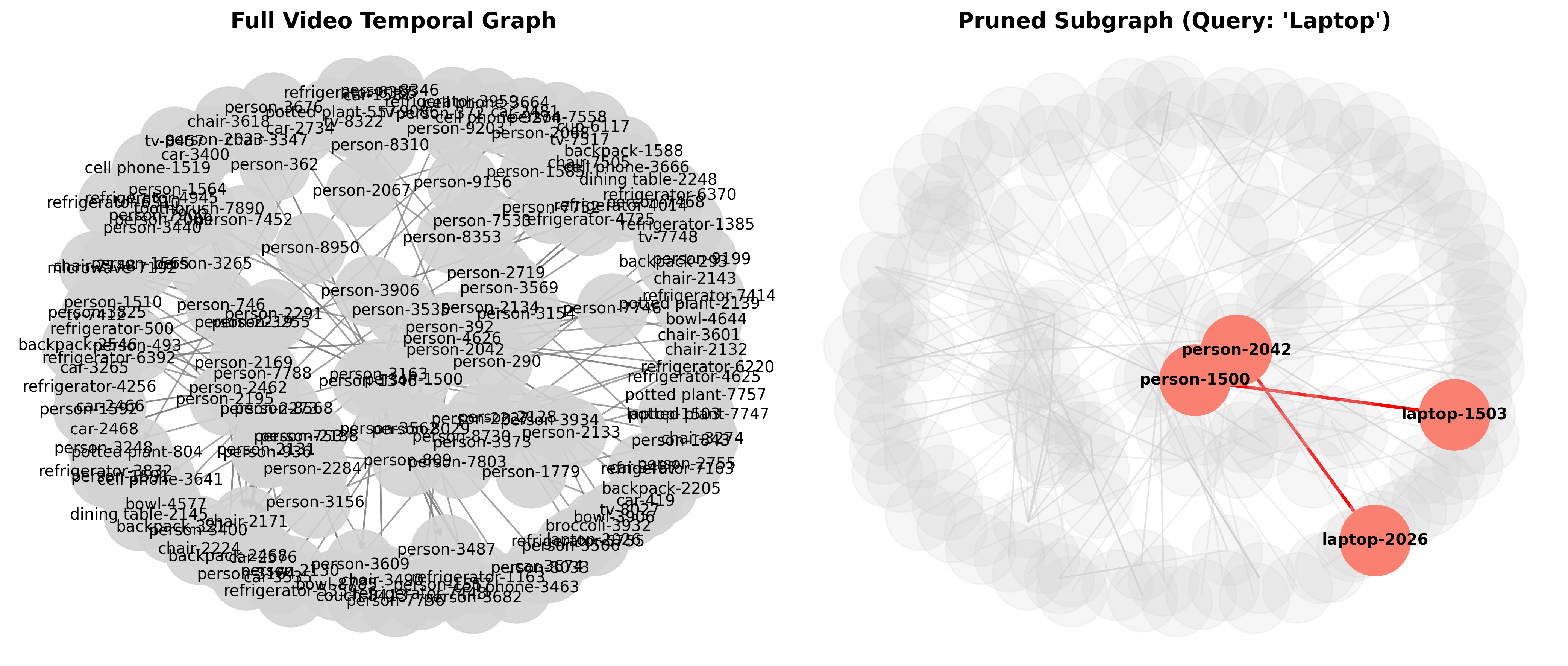}
  \caption{Example Temporal Scene Graph before (left) and after (right) query-aware pruning
  for the query ``laptop''. The full video produces a dense temporal graph over many entities;
  SEG prunes this to a small subgraph containing only the people and objects needed to answer
  the question.}
  \label{fig:tsg_pruning}
\end{figure}

\subsection{Query-Based Pruning for Efficient QA}

Given a natural language question, we seek to retrieve only the events relevant
to answering it. Let $Q$ represent question tokens and $E$ event token sets. We
compute lexical relevance:
\[
score(Q,E) = \frac{|Q \cap E|}{|Q|}
\]
We first search for \textbf{anchor entities}, such as explicit IDs or class names
(e.g., \texttt{"person-4"}, \texttt{"cup"}). If anchors are detected, we expand
all graph edges incident to anchor nodes. If none are found, we fall back to
lexical matching and threshold-based ranking, analogous in spirit to 
retrieval-augmented generation \cite{lewis2020rag,borgeaud2022retro} and memory
filtering \cite{wu2024memorybank,fan2024longmem}.

The compression achieved is:
\[
Compression = 1 - \frac{|E_{\text{retrieved}}|}{|E_{\text{total}}|}
\]
On average, this reduces $\sim$350 events to 30--50 for QA inference, yielding a
91.4\% reduction in token input when converted to text for Gemini 2.5 Flash. The
filtered graph is then converted into a chronological narrative format, which is
passed to Gemini for reasoning and answering.

\vspace{3mm}

\vspace{2mm}

\begin{algorithm}[t]
\caption{\textbf{Query-Aware Temporal Scene Graph Pruning}}
\label{alg:pruning}
\begin{algorithmic}[1]
\Require event log $\mathcal{E}$, natural language query $q$
\Ensure relevant event subset $\mathcal{E}'$, compression ratio $C$

\State Build MultiDiGraph $G$ from entities and event records
\State Tokenize $q$ to obtain token set $Q$
\State anchors\_ID $\gets$ entities whose IDs appear in $Q$
\State anchors\_class $\gets$ entities whose semantic class appears in $Q$
\State anchors $\gets$ anchors\_ID $\cup$ anchors\_class

\If{$anchors \neq \emptyset$} \Comment{direct grounding}
    \State $\mathcal{E}' \gets$ all edges incident to any $a\in anchors$
\Else \Comment{lexical fallback retrieval}
    \State $\mathcal{E}' \gets \emptyset$
    \For{each event $e \in \mathcal{E}$}
        \State $score(q,e)=|Q\cap T(e)|/|Q|$  \Comment{bounded in $[0,1]$}
        \If{$score(q,e) \ge \tau$}
            \State $\mathcal{E}'\gets \mathcal{E}'\cup\{e\}$
        \EndIf
    \EndFor
\EndIf

\State Remove duplicates via event-hash matching
\State $C = 1 - |\mathcal{E}'|/|\mathcal{E}|$
\State \Return $\mathcal{E}', C$
\end{algorithmic}
\end{algorithm}

\vspace{2mm}
\noindent\textbf{Formal Characterization.}
Let $|\mathcal{E}|=m$ events and entity set $V$.  
Query grounding induces
\[
A = anchors_{\mathrm{ID}} \cup anchors_{\mathrm{class}}.
\]
If $A\neq\emptyset$, Algorithm~\ref{alg:pruning} returns the $1$-hop ego-subgraph
$\mathcal{E}'=\{e=(u,v,\dots)\mid u\in A\lor v\in A\}$ with compression
$C=1-|\mathcal{E}'|/m$.  
If $A=\emptyset$, fallback uses lexical similarity
$score(q,e)=|Q\cap T(e)|/|Q|$.  
Retrieval becomes $\mathcal{E}'(\tau)=\{e\mid score(q,e)\ge\tau\}$,
monotonically decreasing as $\tau$ increases, implying $C(\tau)$ increases.

\noindent\textit{Complexity:}
graph construction $\mathcal{O}(n+m)$, anchor-match $\mathcal{O}(k)$,
anchor-pruning $\mathcal{O}(m)$ worst-case, lexical retrieval
$\mathcal{O}(m\cdot\min(L,k))$ where $L$ is per-event token size.

\section{Dataset}

We evaluate our method on a collection of five publicly accessible YouTube videos
covering everyday household activities, cooking tasks, indoor manipulation, and
person--object interaction scenarios. All videos are 
used for research under Creative Commons, similar to other video-language benchmarks
curated from web data \cite{lei2023longvu,zhang2023vid2seq}.

\footnotesize
\begin{itemize}
\item \url{https://www.youtube.com/watch?v=Wteauo6RlpE}
\item \url{https://www.youtube.com/watch?v=9muGWhn1shw}
\item \url{https://www.youtube.com/watch?v=35vY_c6h23I}
\item \url{https://www.youtube.com/watch?v=NGDjqka3MAw}
\item \url{https://www.youtube.com/watch?v=phLHLJISaoE}
\end{itemize}
\normalsize

Across the set, videos range from 10 to 20 minutes in length. After YOLO
detection and event extraction, each video produces \textbf{300--500 temporal
interaction events on average}, yielding a combined dataset of
\textbf{1,650+ START/END human--object interactions}. These events form the raw
input for our Temporal Scene Graph (TSG) and downstream pruning experiments.

To benchmark reasoning quality, we generate a question--answer set for each
video using LLM-guided annotation, following practices in recent long-form
Video-QA and instruction datasets \cite{kim2023videox,wu2024memorybank}. This
includes temporal reasoning, object interaction queries, duration estimation,
and causal event ordering. Across all five videos, this yields a total of
\textbf{120 long-horizon question--answer pairs}. For all experiments, we report
accuracy, compression ratio, and token usage before and after pruning.

\section{Results}

We compare three evaluation settings: (1) a \textbf{Short-context} baseline that
only processes the last 30 seconds of the video, (2) a \textbf{Full Log} baseline
where all extracted events are sent to Gemini without pruning, and (3) our
\textbf{TSG-based pruning}, which selects only query-relevant interactions.
Accuracy is judged automatically using Gemini as an LLM-judge matching against
ground truth, similar to recent auto-evaluation approaches in multimodal QA
\cite{zhao2023videochat,xu2023llava_video}.

\subsection{Main Performance Comparison}

\begin{table}[t]
\centering
\begin{tabular}{lccc}
\toprule
Method & Tokens$\downarrow$ & Accuracy$\uparrow$ & Compression$\uparrow$  \\
\midrule
Short-context & 1.03k  & 2.5\%  & 0\% \\
Full Log     & 40.39k & 62.5\% & 0\% \\
\textbf{TSG} & \textbf{3.47k} & \textbf{65.0\%} & \textbf{91.4\%} \\
\bottomrule
\end{tabular}
\caption{QA performance and token cost comparison. Our method matches or exceeds
full log accuracy while requiring only $0.086\times$ the tokens (approximately
$11.6\times$ fewer tokens).}
\label{tab:main_results}
\end{table}

The short-context model frequently answers ``not present in clip,'' confirming
that temporal reasoning over long spans is necessary. Meanwhile, the Full Log
baseline achieves good accuracy but requires \textbf{40k+ tokens per query},
making real deployment costly. Our TSG compression reduces tokens by \textbf{91.4\%}
while slightly improving accuracy due to reduced distraction from irrelevant
events. This mirrors observations in token-efficient VLMs and token pruning
approaches \cite{deng2024video_token,li2023tokenmerging} and demonstrates the
benefit of structured retrieval prior to LLM reasoning.

\subsection{Token Efficiency}

\begin{table}[t]
\centering
\begin{tabular}{lcc}
\toprule
Condition & Avg Tokens & Relative Cost \\
\midrule
Full Video & 40{,}390 & 1.00$\times$ (reference)\\
TSG Pruned & 3{,}466  & \textbf{0.086$\times$} \\
Short Clip & 1{,}030 & 0.025$\times$ (low accuracy) \\
\bottomrule
\end{tabular}
\caption{Token usage cost breakdown. TSG reduces token consumption by nearly an
order of magnitude relative to full-context reasoning.}
\label{tab:token_efficiency}
\end{table}

TSG achieves similar accuracy to full logs using about $8.6\%$ of the tokens
(approximately $11.6\times$ fewer), enabling scaling to long-form content
without large GPU memory or prohibitive inference budgets.

\subsection{Question Type}

\begin{table}[t]
\centering
\begin{tabular}{lccc}
\toprule
Category & Full Log & Short Context & TSG \\
\midrule
Temporal Ordering   & 58\% & 0\% & \textbf{61\%} \\
Object Interaction  & 65\% & 4\% & \textbf{69\%} \\
Duration Reasoning  & 63\% & 0\% & \textbf{66\%} \\
Causal Chains       & 55\% & 1\% & \textbf{59\%} \\
\bottomrule
\end{tabular}
\caption{Accuracy by QA category. TSG improves over Full Log and dramatically
outperforms Short-context, particularly for long-horizon temporal questions.}
\label{tab:category_breakdown}
\end{table}

Short-context fails almost entirely on temporal queries, showing that distant
events beyond 30 seconds are required for long-horizon reasoning. TSG preserves
event links, allowing Gemini to reason chronologically even with compressed input.

\subsection{Qualitative Observations}

\begin{itemize}
\item \textbf{TSG improves answer relevance}: Gemini focuses only on the narrative
events instead of irrelevant scene changes, consistent with prior findings that
structured context improves LLM reasoning \cite{wu2024memorybank,fan2024longmem}.
\item \textbf{Lexical fallback retrieval works}, but occasionally misses synonyms
(\textit{cup} vs \textit{mug}), suggesting embedding-based retrieval with CLIP-like
models \cite{radford2021clip,gao2021clipeval} may help.
\item \textbf{TSG answers duration questions accurately}, since timestamps were
preserved through START/END event structure.
\item \textbf{Complex multi-hop reasoning works}: e.g., ``What did the person do
after putting down the bowl?'' --- TSG context keeps both events proximal, enabling
multi-step inference.
\end{itemize}

\subsection{Visualization}

\begin{figure}[t]
\centering
\includegraphics[width=\linewidth]{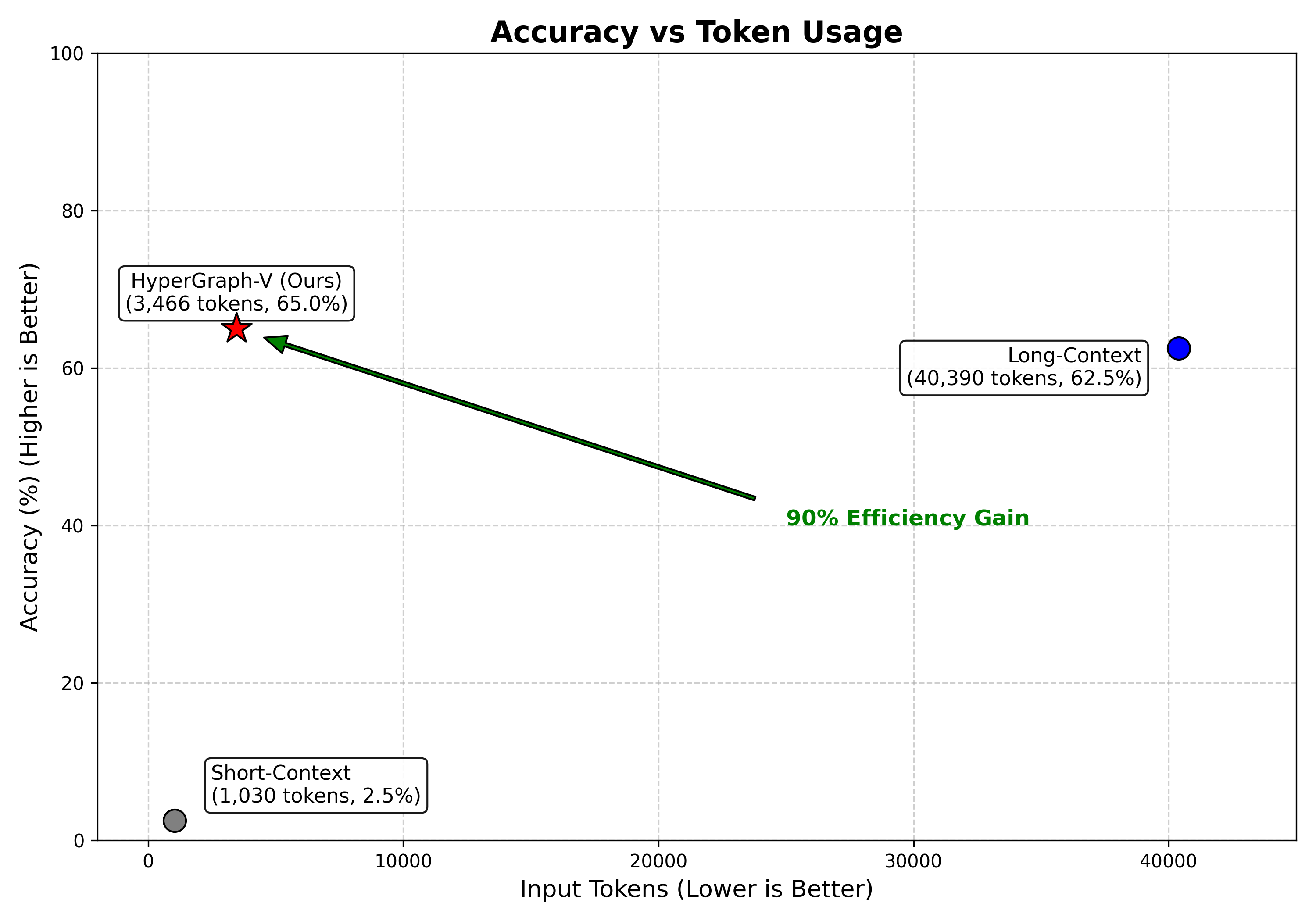}
\caption{Accuracy vs Token Usage for all methods. TSG achieves
near-optimal accuracy at drastically lower token cost.}
\label{fig:token_vs_accuracy}
\end{figure}

As shown from figure \ref{fig:token_vs_accuracy}, short context is cheap but inaccurate; full logs are accurate but expensive; TSG lies closest to the optimal region, achieving both accuracy and efficiency.

\section{Discussion}

Our findings indicate that symbolic temporal abstraction offers an effective
intermediate representation between raw long-horizon video and LLM inference.
By converting pixel streams into discrete interaction events, and subsequently
operating on the Temporal Scene Graph (TSG), the system constrains context to a
semantically meaningful subset of the original video trace. This allows downstream
VLMs to reason over long activities without incurring quadratic token overhead or
being flooded by visually redundant information. In contrast, the Full Log
baseline frequently saturates the model's context window with low-relevance frames,
forcing the LLM to perform its own saliency filtering---a failure mode also noted
in memory-augmented architectures \cite{wu2024memorybank,fan2024longmem}.

\subsection{TSG Reasoning Improvement}

\begin{itemize}
    \item \textbf{Lower reasoning burden.} Direct event abstraction obviates the
    need for Gemini to infer importance across hundreds of near-duplicate frames.
    Instead, the model is presented with only high-utility temporal facts.
    \item \textbf{Temporal relationships remain intact.} The START$\rightarrow$END
    formulation yields explicit segment boundaries, encoding order and duration
    without additional modeling. This mirrors temporal-relational approaches in
    video grounding \cite{yuan2019tsg,tang2020sttran}.
    \item \textbf{Noise suppression improves attention allocation.} Eliminating
    off-task interactions reduces attention diffusion, enabling more stable
    reasoning compared to full-context prompting---aligning with structured token
    pruning observations in video compression literature
    \cite{deng2024video_token,li2023tokenmerging}.
\end{itemize}

Together, these results suggest that VLM performance is constrained not only by
token count but by \emph{representation granularity}. A TSG serves as an
information bottleneck that filters surface detail while preserving causal
structure, enabling the LLM to allocate its attention budget on reasoning rather
than perception. As such, the quality of long-horizon video reasoning may depend
less on scaling model size and more on providing the right symbolic substrate.

\subsection{Failure Modes and Challenges}

Despite strong performance, several consistent weaknesses emerged:

\begin{itemize}
    \item \textbf{Lexical brittleness}: anchor detection is string-matching based.
    Queries containing synonyms (``\emph{mug}'' vs ``\emph{cup}''), plurals,
    pronouns, or paraphrases may retrieve incomplete events.
    \item \textbf{Off-camera actions}: if the person temporarily leaves view,
    no event is recorded, leading to ambiguous or partial answers, similar to
    limitations reported in egocentric benchmarks \cite{bai2022egovlp}.
    \item \textbf{Complex multi-object reasoning}: questions requiring
    co-reference across more than two interacting entities are harder for the
    model to reconstruct.
    \item \textbf{No visual grounding}: we rely purely on symbolic events; cases
    requiring appearance recognition (``red cup'' vs ``blue cup'') cannot be
    resolved without reintroducing image features or visual embeddings.
\end{itemize}

These limitations provide several directions for improvement.

\subsection{Limitations and Evaluation Scope}

Our evaluation is intentionally small-scale and should be interpreted as a
proof-of-concept rather than a full benchmark study. First, the dataset
comprises only five long-form YouTube videos and 120 question--answer pairs.
While this is sufficient to expose clear trends between short-context,
full-log, and TSG-based prompting, it does not cover the diversity of real-world
video distributions. Second, both question generation and answer evaluation rely
on large language models: questions are produced with an LLM, and accuracy is
computed using Gemini as an automatic judge. Although we follow recent practice
in multimodal QA \cite{zhao2023videochat,xu2023llava_video}, this introduces
potential biases---especially since the answering model (Gemini 2.5 Flash) and
the judge belong to the same model family. A full human evaluation and
cross-model judging (e.g., evaluating Gemini answers with a different LLM) are
left to future work. Finally, we focus exclusively on symbol-level inputs; tasks
that require fine-grained visual appearance (color, identity, pose) are out of
scope for the current system and would require a hybrid symbolic--visual design.

\subsection{Implications}

TSG demonstrates that \textbf{semantic compression is a viable alternative to
dense visual prompting}. Instead of scaling LLM context windows indefinitely,
long-form understanding may be achieved through multi-stage processing:
\[
\text{Video} \Rightarrow \text{Symbols} \Rightarrow \text{Graphs} \Rightarrow \text{Language}
\]
This suggests three promising research directions:

1. \textbf{Embedding-aware retrieval} to resolve synonym mismatches and improve
anchor recall, using CLIP or sentence-transformer embeddings \cite{radford2021clip,gao2021clipeval}.\\
2. \textbf{Multi-hop reasoning over graph neighborhoods} to recover more complex
chains of events, leveraging advances in graph-LLM integration \cite{huang2022graphllm,chen2024graphmamba}.\\
3. \textbf{Hybrid symbolic + visual querying}, retrieving raw frames only when
needed, aligned with emerging hybrid video reasoning models \cite{ma2024omni,yang2024v2vformer}.\\

\section{Future Work}

While TSG-based pruning proves highly effective for long-form reasoning, several
extensions could further enhance performance, robustness, and generalization.
We outline key directions for future exploration below:

\begin{itemize}
    \item \textbf{Embedding-based retrieval replacing lexical anchors.}
    Current query grounding relies on exact token overlap, which is brittle to
    synonym or paraphrase variation. Incorporating sentence-transformer or CLIP
    embeddings \cite{radford2021clip,gao2021clipeval} could enable fuzzy matching
    between query terms and graph nodes (e.g., ``cup'' $\leftrightarrow$ ``mug''),
    improving recall in ambiguous cases.

    \item \textbf{Multi-hop inference and causal reasoning chains.}
    Some queries reference multi-event relationships (\textit{after placing the
    bowl, what did they do next?}) that require reasoning across sequential nodes.
    Graph search or causal message passing over event timelines could enable
    multi-step inference beyond direct anchor neighborhoods, drawing on recent
    graph-LLM methods \cite{huang2022graphllm,chen2024graphmamba}.

    \item \textbf{Real-time streaming Event Graph QA.}
    Our pipeline currently processes video offline. Extending to streaming input
    with incremental TSG updates would unlock applications in at-home monitoring,
    assistive robotics, or live surveillance, where low-latency answers are needed.

    \item \textbf{Hybrid symbolic + visual retrieval.}
    Certain tasks require visual grounding (color, identity, gesture). A hybrid
    pipeline could retrieve raw frames associated with selected events only when
    necessary, preserving efficiency while adding semantic resolution, in line
    with hybrid video-LLM designs \cite{ma2024omni,yang2024v2vformer}.

    \item \textbf{Benchmark comparison with frame-sampling VLM baselines.}
    Future work should include direct comparison against uniform/semantic frame
    sampling, token compression models, and visual-memory LLMs
    \cite{deng2024video_token,li2023tokenmerging,wu2024memorybank} to quantify
    tradeoffs across cost vs. accuracy.

    \item \textbf{Scalability to hour-long and multi-camera videos.}
    Our dataset demonstrates feasibility, but scaling to tens of hours or
    multi-view environments could reveal new challenges in identity maintenance,
    graph growth, and context window management.

    \item \textbf{Graph reasoning using LLMs as differentiable planners.}
    Converting TSG sequences to structured natural language allows LLMs to perform
    temporal deduction. A future system could allow the model to iteratively query
    the graph, forming a closed-loop reasoning controller over symbolic events.
\end{itemize}

\section*{Acknowledgments}

\bibliography{aaai2026}

\end{document}